\documentclass[letterpaper]{article} % DO NOT CHANGE THIS
\usepackage{aaai25}  % DO NOT CHANGE THIS
\usepackage{times}  % DO NOT CHANGE THIS
\usepackage{helvet}  % DO NOT CHANGE THIS
\usepackage{courier}  % DO NOT CHANGE THIS
\usepackage[hyphens]{url}  % DO NOT CHANGE THIS
\usepackage{graphicx} % DO NOT CHANGE THIS
\usepackage{natbib}  % DO NOT CHANGE THIS AND DO NOT ADD ANY OPTIONS TO IT
\usepackage{caption} % DO NOT CHANGE THIS AND DO NOT ADD ANY OPTIONS TO IT
\usepackage{algorithm}
\usepackage{algorithmic}

\usepackage{newfloat}
\usepackage{listings}
\DeclareCaptionStyle{ruled}{labelfont=normalfont,labelsep=colon,strut=off} % DO NOT CHANGE THIS
\floatstyle{ruled}
\newfloat{listing}{tb}{lst}{}
\floatname{listing}{Listing}

\usepackage{url}            % simple URL typesetting
\usepackage{booktabs}       % professional-quality tables
\usepackage{microtype}      % microtypography
\usepackage{xcolor}         % colors
\usepackage{graphicx}
\usepackage{todonotes}
\usepackage{makecell}
\usepackage{subcaption}
\usepackage{boldline}
\usepackage{amsmath}
\usepackage{dblfloatfix}

\newcommand{\OurMethod}{LCP}

\title{Representing Positional Information in\\ Generative World Models for Object Manipulation}

\author{
  Stefano Ferraro$^1$ \qquad
  Pietro Mazzaglia$^1$ \qquad
  Tim Verbelen$^2$ \qquad
  Bart Dhoedt$^1$ \qquad
  Sai Rajeswar$^3$ 
  \vspace{0.25cm} \\
}

\affiliations{  $^1$ Ghent University, $^2$ VERSES, $^3$ ServiceNow\\ 
  stefano.ferraro@ugent.be}
  
\begin{document}

\maketitle

\begin{abstract}
Object manipulation capabilities are essential skills that set apart embodied agents engaging with the world, especially in the realm of robotics. The ability to predict outcomes of interactions with objects is paramount in this setting. While model-based control methods have started to be employed for tackling manipulation tasks, they have faced challenges in accurately manipulating objects. As we analyze the causes of this limitation, we identify the cause of underperformance in the way current world models represent crucial positional information, especially about the target's goal specification for object positioning tasks. 
We introduce a general approach that empowers world model-based agents to effectively solve object-positioning tasks. We propose two declinations of this approach for generative world models: position-conditioned (PCP) and latent-conditioned (LCP) policy learning. In particular, LCP employs object-centric latent representations that explicitly capture object positional information for goal specification. This naturally leads to the emergence of multimodal capabilities, enabling the specification of goals through spatial coordinates or a visual goal. Our methods are rigorously evaluated across several manipulation environments, showing favorable performance compared to current model-based control approaches. 
% ations, we introduce a novel technique that augments object-centric world models with a representation that explicitly captures object positional information. 
\end{abstract}

%
% 2 - Preliminaries
% 2.1 - World Models
% 2.2 - Object positional tasks
% 3 - Limitations of world model for object positional tasks (analysis of performance and reconstructions) -> at the end we can describe some potential naive solutions, and already say that they have better performance but they are not ideal/definitive/solid
% 4 - Skill FOCUS (or a moe general Solutions chapter)
% 5 - Experiments
% 

\section{Introduction}

% \begin{wrapfigure}{r}{5cm}
%     \centering
%     \includegraphics[width=5cm]{figures/teaser.png}
%     \caption{Object positioning task}
%     \label{fig:teaser}
% \end{wrapfigure}

For robotic manipulators, replicating the tasks that humans perform is extremely challenging, due to the complex interactions between the agent and the environment. In recent years, deep reinforcement learning (RL) has emerged as a promising approach for addressing these challenging scenarios \cite{Levine2016DeepVMPolicies, OpenAI2019Rubik, Kalashnikov2018QTOpt, Lu2021AWOpt, Lee2021Stacking}. % ferraro2022computational
Among RL algorithms, model-based approaches aim to provide greater data efficiency compared to their model-free counterparts \cite{Fujimoto2018TD3, Haarnoja2018SAC}. With the advent of world models (WM) \cite{Ha2018WM}, model-based agents have demonstrated impressive performance across various domains \cite{Hafner2020DreamerV1, Rajeswar2023MasteringURLB, Hafner2023DreamerV3, Hansen2022TDMPC, hansen2024tdmpc2, lancaster2024modemv2}, including real-world robotic applications \cite{Wu2022DayDreamer, Seo2023MVMWM}.

% intro to object centric world models, and what do they solve
When considering object manipulation tasks, it seems natural to consider an object-centric approach to world modeling. Object-centric world models, like FOCUS \cite{ferraro2023focus} learn a distinct dynamical latent representation \textbf{per object}. This contrasts with the popular Dreamer method \cite{Hafner2023DreamerV3}, where a single \textbf{flat} representation, referring to the whole scene is extracted.   

\begin{figure}[!tbp]
    \centering
    \includegraphics[width=0.8\linewidth]{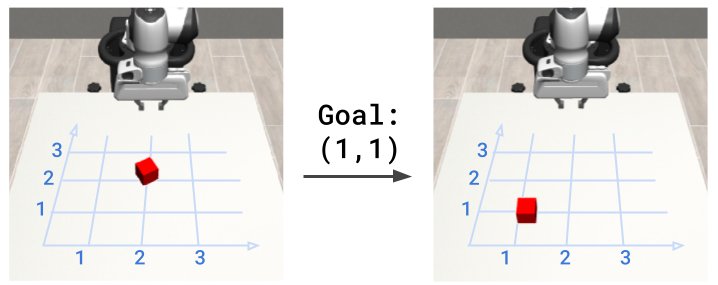}
    \caption{Object positioning task with coordinates goal.}
    \label{fig:obj_positioning}
\end{figure}
Model-based generative agents, like Dreamer and FOCUS, learn a latent model of the environment dynamics by reconstructing the agent's observations and use it to generate latent sequences for learning a behavior policy in imagination \cite{Hafner2020DreamerV1, Hafner2021DreamerV2, Hafner2023DreamerV3}. However, these kinds of agents have shown consistent issues in succeeding in object manipulation tasks, both from proprioceptive/vector inputs \cite{hansen2024tdmpc2} and from images \cite{Seo2022MaskWM}. For instance, in the \textit{Shelf Place} task from Metaworld \cite{Yu2019MetaWorld}, DreamerV3's success rate after 2M steps oscillates around 25\% when other approaches, such as the simpler SAC baseline \cite{Haarnoja2018SAC}, quickly reach 100\% success, as reported in \cite{hansen2024tdmpc2}. 

In this work, we aim to study this failure mode of generative model-based agents and propose a set of solutions that alleviate these issues in existing world models. To be able to perform controlled experiments, in a simple but widely used setup, we focus on the \textit{visual object positioning} problem. As shown in Figure \ref{fig:obj_positioning}, in this setting, at each timestep, the agent receives a visual observation, which shows what the current scene looks like, and a goal target, representing where an object should be moved. Notably, the target could be expressed either as a vector, containing the spatial coordinates of the target location or as an image, showing the object in the desired pose and location. 
% containing the target location where to move one of the objects in the scene

% First, we observe that even in simple settings, where the object is attached to the agent, generative agents suffer. 
After analyzing the causes of failure in this setup, we propose two solutions to improve performance: a simpler solution that applies to any world model architecture and a more tailored solution for object-centric world models. Finally, we evaluate the proposed methods over a set of object positioning tasks, showing a neat improvement over standard methods.
To summarize, the contributions of this work are as follows:
\begin{itemize}
    \item an analysis of the reasons for the inefficiency of generative model-based agents for solving tasks that require positional information;
    \item a simpler solution that presents no major changes to the world model architecture and minimal changes to policy learning. This solution already strongly improves performance in several tasks, where the target is expressed as a vector of spatial coordinates;
    \item a tailored solution employing an object-centric approach that integrates positional information about the objects into the latent space of the world model. This approach enables the possibility to specify goals through multimodal targets, e.g. vector inputs or visual goals. 
\end{itemize}

\section{Preliminaries}
% Setup and condition considered.
The agent is a robotic manipulator that, at each discrete timestep $t$ receives an input $x_t$ from the environment. The goal of the agent is to move an object in the environment from its current position $p_t^{obj}$ to a target goal position $p_g^{obj}$. 

In this work, we focus on observations composed of both visual and vector entities. Thus, $x_t = (o_t, v_t)$ is composed of the visual component $o_t$ and of the vector $v_t$. The latter is a concatenation of proprioceptive information of the robotic manipulator $q_t$, the object's position $p_t^{obj}$, and the target position $p_g^{obj}$. The target position can also be expressed through a visual observation $x_g$, from which the agent should infer the corresponding $p_g^{obj}$ to succeed in the positioning task.
% \begin{align*}
% \begin{split}
% x_t = (o_t, v_t), & \qquad 
% v_t = (q_t, p_t^{obj}, p_g^{obj})
% \end{split}
% \end{align*}

\subsection{Generative World Models}
\label{sec:wm}
% intro on world models, mention on modification on DreamerV2 to match it with DreamerV3

Generative world models learn a latent representation of the agent inputs using a variational auto-encoding framework \cite{kingma2022autoencoding}. Dreamer-like agents \cite{Hafner2021DreamerV2, Hafner2023DreamerV3} implement the world model as a Recurrent State-Space Model (RSSM) \cite{hafner2019planet}. The encoder $f(\cdot)$ is instantiated as the concatenation of the outputs of a CNN for high-dimensional observations and an MLP for low-dimensional proprioception. Through the encoder network, the input $x_t$ is mapped to an embedding $e_t$, which then is integrated with dynamical information with respect to the previous RSSM state and the action taken $a_t$, resulting in $s_t$ features. 
\begin{align*}
\begin{split}
\textrm{Encoder:} & \quad e_t = f(x_t) \\ 
\textrm{Posterior:} & \quad p_{\phi}(s_{t+1}|s_t, a_t, e_{t+1}), \\ 
\textrm{Prior:} & \quad p_{\phi}(s_{t+1}|s_t, a_t), \\
% \textrm{Reward predictor:} & \quad p_{\theta}(\hat{r}_t|s_t), \\
\textrm{Decoder:} & \quad p_{\theta}(\hat{x}_t|s_t). \\
\end{split}
\end{align*}
Generally, the system either learns to predict the expected reward given the latent features \cite{Hafner2020DreamerV1}, using a reward predictor $p_{\theta}(\hat{r}_t|s_t)$. Alternatively, some world-model based methods adopt specialized ways to compute rewards in imagination, as the goal-conditioned objectives in LEXA \cite{mendonca2021discovering}. 

 Rewards are computed on rollouts of latent states generated by the model and are used to learn the policy $\pi$ and value network $v$ in imagination \cite{Hafner2020DreamerV1, Hafner2021DreamerV2, Hafner2023DreamerV3}.

In our experiments, we consider a world model with a discrete latent space \cite{Hafner2021DreamerV2}. We also implement advancements of the world model representation introduced in DreamerV3 \cite{Hafner2023DreamerV3}, such as the application of the $\textrm{symlog}$ transform to the inputs, KL balancing, and free bits to improve the predictions of the vector inputs and the robustness of the model.

% A PyTorch implementation of DreamerV2 is used. Relevant changes of DreamerV3 have been integrated in the codebase for a fair comparison to the state of the art. The symlog function is introduced for the scaling of the state inputs. KL balancing and free bits are integrated for a robust trade off between representation learning and latent dynamics. \todo{details about symlog and kl balancing}
\begin{figure}[!t]

    \centering
    \includegraphics[width=0.9\linewidth]{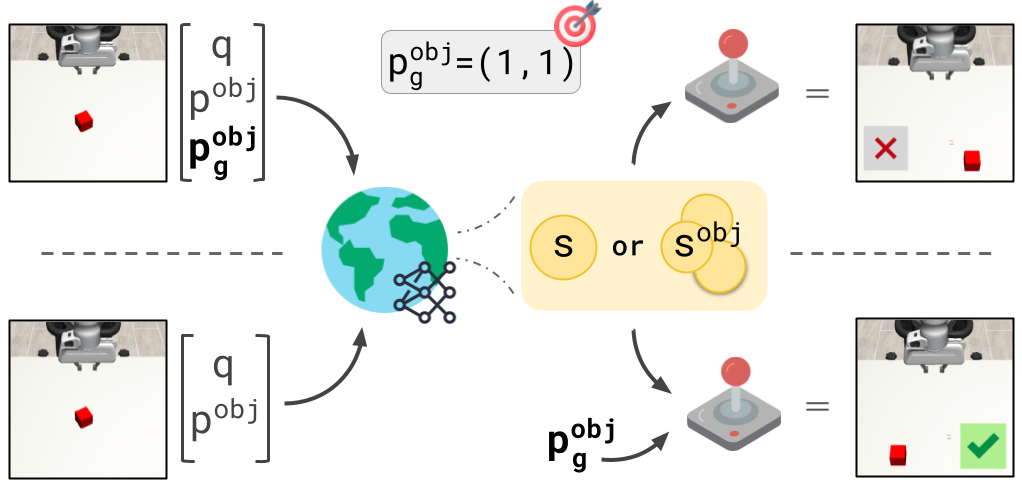}

    % \centering
    % \begin{subfigure}[t]{0.3\textwidth}
    %     \centering
    %     \includegraphics[height=1.75in]{figures/ocwm_2.png}
    %     \caption{Object-centric WM + Policy}
    % \end{subfigure}%
    % \centering
    % \begin{subfigure}[t]{0.3\textwidth}
    %     \centering
    %     \captionsetup{justification=centering}
    %     \includegraphics[height=1.75in]{figures/tc-ocwm_2.png}
    %     \caption{Flat/Object-centric WM + goal-conditioned Policy}
    % \end{subfigure}%
    
    % \caption{\textbf{(a)} The \textbf{world model} compresses visual observations and state vector into a single latent state representation. Crucially, the compressed representation serve as input to the policy for action selection. Target information is provided through the input state vector. \textbf{(b)}: In the case of an \textbf{object-centric world model} the latent representation is composed of distinct latent states per object. \textbf{(c)}: Both single and object-centric representations can be paired to a \textbf{target-conditioned policy}.}

    \caption{ The \textbf{world model} compresses visual observations and state vector into a latent state representation. Crucially, the compressed representation serves as input to the policy for action selection. The world model can either be \textbf{flat}, encoding a single latent state, or \textbf{object-centric}, where the latent representation consists of distinct latent states for each object. \textbf{(top)} Goal information is provided through the input state vector. \textbf{(bottom)}: Both single and object-centric representations can be paired to a \textbf{target-conditioned policy}.}
    \label{fig:wms}
\end{figure}

\subsection{Object-centric World Models}
\label{sec:ocwm}

Compared to Dreamer-like \textit{flat} world models, the world model of FOCUS \cite{ferraro2023focus} introduces the following object-centric components:
\begin{align*}
\begin{split}
% \textrm{Decoder:} & \quad p_{\theta}(\hat{v}_t^{scene}|s_t), \\
\textrm{Object latent extractor:} & \quad p_{\theta}(s_t^{obj}|s_t, c^{obj}), \\
% \textrm{Position encoder:} & \quad \hat{s_t}^{obj} = f_{obj}(p_t^{obj}), \\
\textrm{Object decoder:} & \quad p_{\theta}(\hat{x}_t^{obj}, \hat{m}_t^{obj}|s_t^{obj}).
\end{split}
\end{align*}
% \vspace{3pt}

% $v_t^{scene}$ represents here all the scene inputs, that are not related to objects, and it is composed of proprioceptive information $q_t$ and the target positions $p_T^{obj}$. 
% $$v_t^{scene} = (q_t, p_T^{obj})$$
Here, $x_t^{obj} = (o_t^{obj}, p_t^{obj})$ represents the object-centric inputs and it is composed of segmented RGB images $o_t^{obj}$ and object positions $p_t^{obj}$. The variable $c^{obj}$ indicates which object is being considered.
% $$x_t^{obj} = $$

Thanks to the \textit{object latent extractor} unit, object-specific information is separated into distinct latent representations $s_t^{obj}$.
Two decoding units are present. The introduced object-centric decoder $p_{\theta}(\hat{x}_t^{obj}, \hat{m}_t^{obj}|s_t^{obj})$ reconstructs each object's related inputs $x_t^{obj}$ and segmentation mask $m_t^{obj}$. The original Dreamer-like decoder takes care of the reconstruction of the remaining vector inputs, i.e. proprioception $q_t$ and given goal targets $p_g^{obj}$.  

We provide additional descriptions of the world model and policy learning losses, hyperparameters, and training details in the Appendix. 

% Training of the FOCUS architecture is guided by following loss function:
% \begin{equation}
% \mathcal{L_\textrm{FOCUS}} = \mathcal{L}_\textrm{dyn} + \mathcal{L}_\textrm{state} + \mathcal{L}_\textrm{obj}.
% \end{equation}
% $\mathcal{L}_\textrm{dyn}$ refers to the dynamic component of the RSSM, and equals too:
% \begin{equation}
%     \mathcal{L_\textrm{dyn}} = D_\textrm{KL}[p_{\phi}(s_{t+1}|s_t, a_t, e_{t+1}) || p_{\phi}(s_{t+1}|s_t, a_t)].
% \end{equation}
% the backpropagation is balanced and clipped below 1 nat as in DreamerV3 \cite{Hafner2023DreamerV3}.

% The decoder learns to reconstruct the vector state information by $\hat{v}_t$ by minimization of the negative log-likelihood (NLL) loss:
% \begin{equation}
% \begin{split}
%     \mathcal{L}_\textrm{state} = -\log p_\theta(\hat{v}_t^{scene}|s_t)
% \end{split}
% \end{equation} 

% Finally, the object loss component is instantiated as the composition of NLL over the mask and RGB mask reconstructions:
% \begin{equation}
% \begin{split}
%     \mathcal{L_\textrm{obj}} = -\log \underbrace{p(\hat{m_t})}_\textrm{mask}  -\log \sum^{N}_{\textrm{obj}=0} \underbrace{m_t^\textrm{obj} p_{\theta}(\hat{x_t^\textrm{obj}}|s_t^\textrm{obj})}_\textrm{masked reconstruction}
% \end{split}
% \end{equation}

\subsection{Object Positioning Tasks}
\label{sec:pos_tasks}

In general terms, we consider positioning tasks the ones where an entity of interest has to be moved to a specific location. Two positioning scenarios are considered in this analysis: \textit{pose reaching} and \textit{object positioning}. Pose-reaching tasks can be seen as simplified positioning tasks where the entity of interest is part of the robotic manipulator itself. Pose-reaching tasks are interesting because these only require the agent to have knowledge of the proprioceptive information to infer their position in space and reach a given target.
When interacting with objects instead, there is the additional necessity of knowing the position of the object entity in the environment. Then, the agent needs to be able to manipulate and move the entity to the provided target location.
 
% visual inputs with targets that are specified via numeric value
For object positioning tasks, especially when considering a real-world setup, there is a significant advantage in relying mainly on visual inputs. It is convenient because it avoids the cost and difficulty associated with tracking additional state features, such as the geometrical shape of objects in the scene or the presence of obstacles.  
Some synthetic benchmarks additionally make use of "virtual" visual targets for positioning tasks \cite{tunyasuvunakool2020dmc, Yu2019MetaWorld}, which strongly facilitates the learning of these tasks, leveraging rendering in simulation. However, applying such "virtual" targets in real-world settings is not often feasible. Non-visual target locations can be provided as spatial coordinates. Alternatively, an image showing the target location could be used to specify the target's position.
% or, thanks to the advent of LLMs, as descriptive information (e.g. "place the red object on the table at the right of the blue object"). In this work, we consider numerical coordinates as target definition.

% cite work for intricate reward definitions to motivate our choice
\textbf{Rewards and evaluation criteria.} When applying RL algorithms to a problem, a heavily engineered reward function is generally necessary to guide the agent's learning toward the solution of the task \cite{OpenAI2019Rubik}. 
% Defining such reward signal can be challenging \cite{}. Moreover, detailed dense rewards require often many sensory signals from the environment (e.g. contact with object, velocities of objects , etc.), which is often an issue when having to set up a reinforcement learning algorithm in the real world. 
The object positioning setup allows us to consider a natural and intuitive reward definition that scales across different agents and environments. We define the reward as the negative distance between the position of the entity of interest and the goal target position:
\begin{equation}
r_t = -\textrm{\texttt{distance}}(\textrm{object}, \textrm{target}) = -\lVert{p^{obj}_t - p^{obj}_g}\lVert_2.
\label{eq:dist}
\end{equation}

In the spirit of maintaining a setup that is as close as possible to a real-world one, to retrieve positional information $p_t$ of the objects we rely on image segmentation information, rather than using the readings provided from the simulator. For each entity of interest, the related position is extracted by computing the centroid of the segmentation mask and subsequently transformed according to the camera extrinsic and intrinsic matrices to obtain the absolute position with respect to the workspace.

For evaluation purposes, we use the goal-normalized score function:
\begin{equation}
% \textrm{\texttt{score\_fn}} = \exp{ ( - \frac{\lVert{p_t - p_T}\lVert}{\lVert{p_T}\lVert} ) }
\textrm{\texttt{normalized score}} = \exp{ \bigg( - \frac{\lVert{p^{obj}_t - p^{obj}_g}\lVert_2}{\lVert{p^{obj}_g}\lVert_2 }\bigg) }
\label{eq:score_fn}
\end{equation}
As detailed in the Appendix, the above function allows us to rescale performance between 0 and 1, where $1$ = expert performance, a common evaluation strategy in RL \cite{cobbe2020procgen, fan2023atarireview}.

% When the relative distance between the entity and target is large the negative exponential will approximate to $0$, while for a small relative distance the score will tend to $1$. 

% This setup has shown good performance as robustness in extracting the positional information of the entity of interest without any major computational overhead. 
% Moreover, such a pipeline would also work in a real-world scenario without the need for extra specialized sensors.
\begin{figure}[!t]
    \centering
    % \begin{figure}[t]{\linewidth}
    %     \centering
    \includegraphics[width=\linewidth]{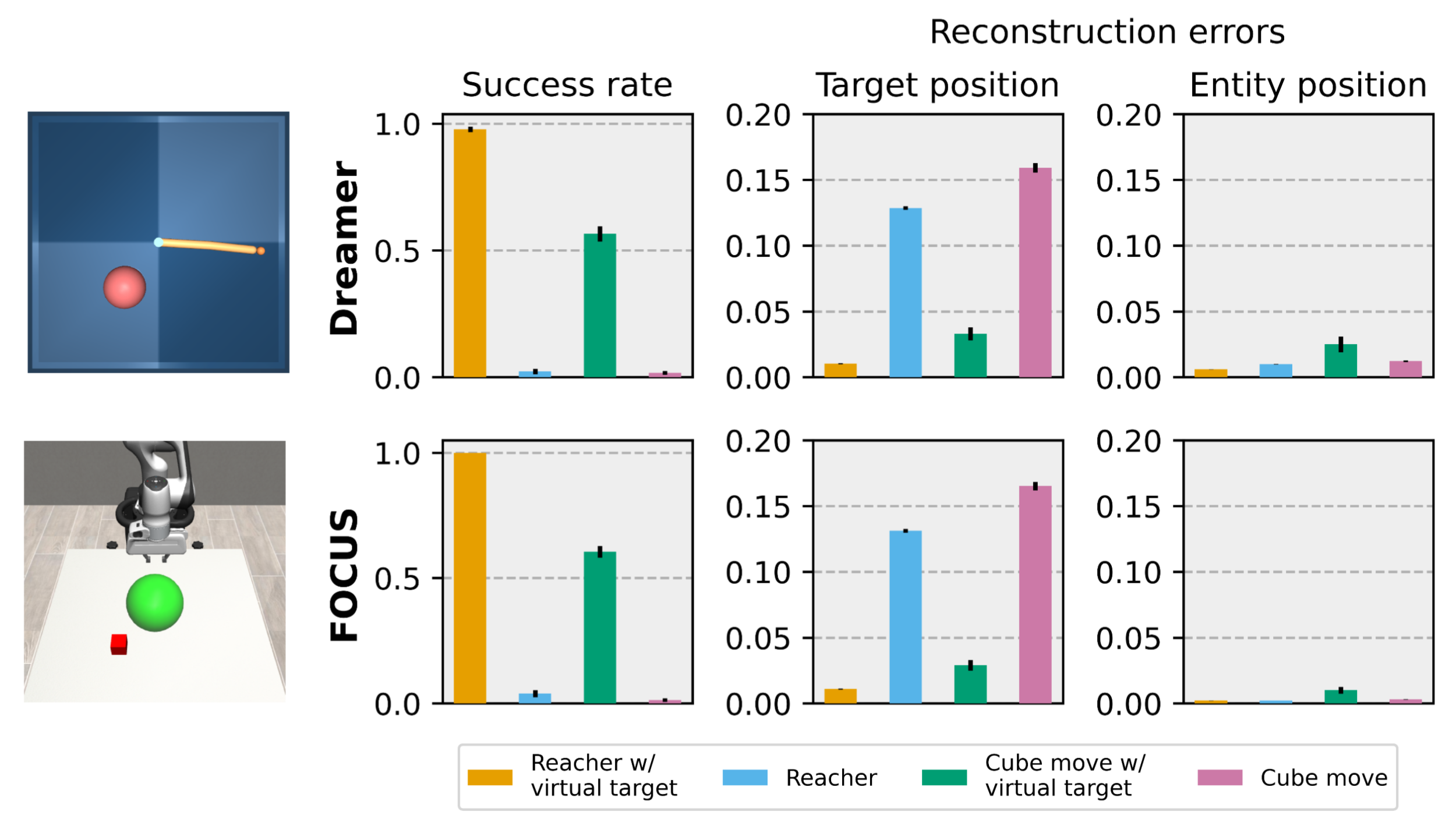}
    % ~ 
    % \centering
    % \begin{subfigure}[t]{0.5\textwidth}
    %     \centering
    %     \includegraphics[height=1.2in]{figures/rec_focus.png}
    %     \caption{}
    % \end{subfigure}%
    
    \caption{\textbf{left}: examples of virtual targets visualization.\textbf{top-right}: Dreamer's success rate and reconstruction performance over target and entity position (end-effector position for reacher and cube position for the cube move environment). \textbf{bottom-right}: Equivalent for the FOCUS object-centric model.
    The success rate for both environments is defined as the entity of interest being within 5cm from the given target at the termination of the episode. Reconstruction errors are computed as L2-norm. }
    \label{fig:reconstructions}
\end{figure}

\section{Analysis of the Current Limitations}
\label{sec:limitations}

To provide insights into the limitations of current world model-based agents in object-positioning tasks, we consider the performance of Dreamer and FOCUS on a pose-reaching and an object-positioning task.

For pose-reaching, we opted for the Reacher environment from the DMC suite \cite{tunyasuvunakool2020dmc}. In this task, we consider the end-effector of the manipulator as the entity to be positioned at the target location. For the more complex object positioning task, we opted for a cube-manipulation task from Robosuite \cite{Zhu2020Robosuite}. The given cube has to be placed at the specified target location to succeed in the task. 

In both environments, the target position is uniformly sampled within the workspace at every new episode. We test the environments in two different scenarios: first, with a virtual visual target that is rendered in the environment, and second, without a visual target, where the target location is provided only as a vector in the agent's inputs.

Dreamer and FOCUS agents are trained for 250k gradient steps on both environments in an offline RL fashion. Datasets of 1M steps are collected adopting the exploration strategy presented in \cite{ferraro2023focus}. Both the world model and the policy are updated at every training step. In Figure~\ref{fig:reconstructions}, we present an overview of the agents' success rate and of the capabilities at reconstructing positional information.

We make the following observations: 
\begin{itemize}
    \item The agents' performance is comparable, with FOCUS being slightly more successful and more accurate in predicting the target and entity position. This is likely thanks to the object-centric nature of the approach;
    \item There is a significant gap in performance between the tasks with the virtual visual targets rendered in the environment and the tasks using only spatial coordinates as a target. The agents struggle to solve the tasks without a virtual target;
    \item There is a negative correlation between the agents' ability to reconstruct positional information and the performance on the task. The lower the reconstruction errors, the higher the success rate on the task. This is particularly evident for the target position, but it also seems to apply to the entity position, i.e. might be the reason for FOCUS performing better than Dreamer.
\end{itemize}
In the remainder of this section, we aim to provide an explanation for the above observations.

% Only in the case of the reach task where the visual target is present the two models have a high success rate. If we consider the other tasks, for both Dreamer and FOCUS the reconstruction in the entity position is adequate while the error in target reconstruction is x10 bigger than the reconstruction error with the visual target present. Ultimately, when lacking a visual cue for the target, both models are able to infer the position of the object in the scene but fails at reconstructing the target position.

\begin{figure}[!b]
    \centering
    \includegraphics[width=0.9\linewidth]{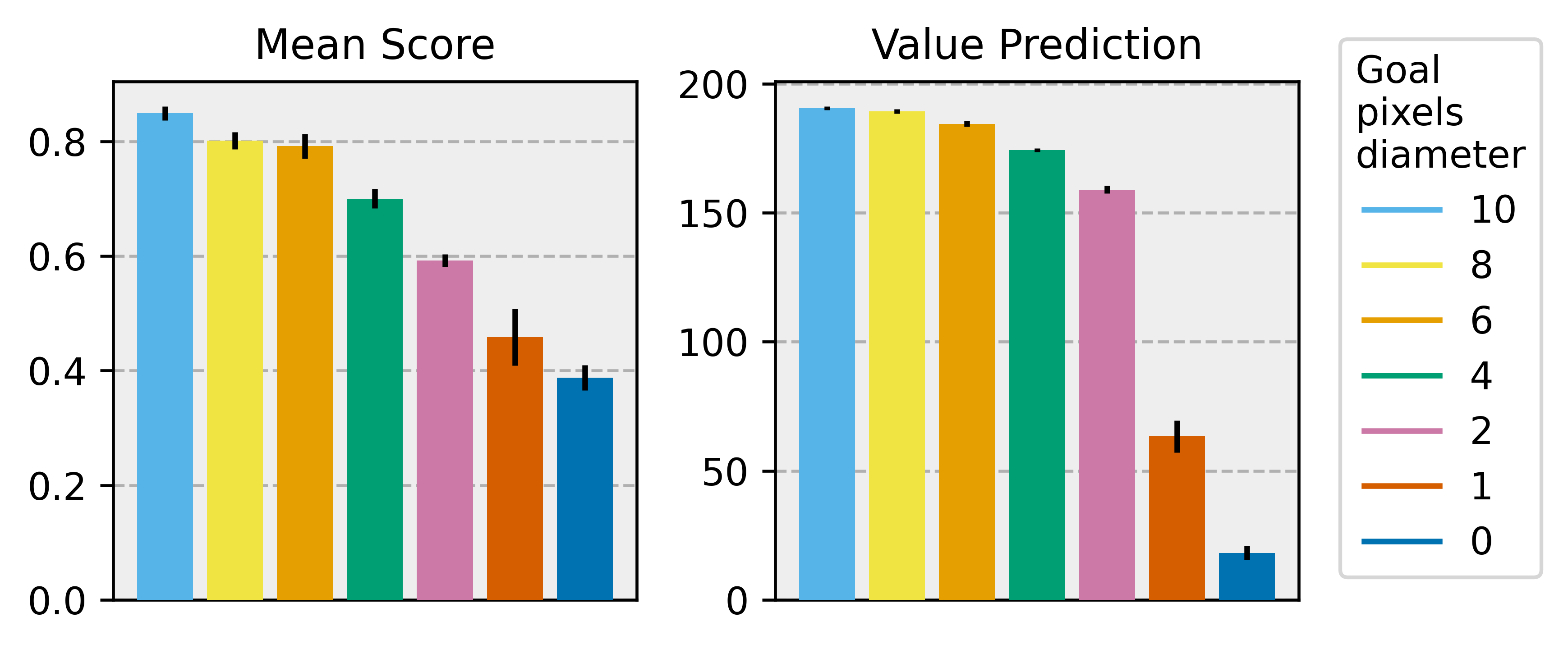}
    \caption{Dreamer virtual visual goal modulation experiments on the Reacher environment. Value prediction from the value network is shown to highlight the policy's awareness of the lack of information with respect to the target goal.}
    \label{fig:target_ablation}
\end{figure}

\textbf{Why is the visual cue so important?} There is a significant difference in the relative significance of the target information compared to the entire observation, in terms of their dimensionality. The information pertaining to a positional target comprises a maximum of three values (i.e., the \textit{xyz} coordinates of the target). Conversely, when considering a visual cue, there are three values (i.e., RGB values) for each pixel that represents the target cue. Consequently, the relative significance of the target information is, at least, greater in the case of a large visual target, i.e. larger than a single pixel. This difference in the dimensionality affects the computation of the loss, and thus the weight of each component in the decoder's loss.
% why it does not happen in case of the entity position?
For the entity, the agents have access to this information in the visual observation. Indeed, it's not surprising that both agents reconstruct the entity position accurately. 

% The results presented in Fig. \ref{fig:reconstructions}.b, have a similar trend to the Dreamer one. The extra object's positional information provided as object masking to the FOCUS model, does not make a difference in the overall performance of the system in absence of a visual target.

% In both cases, inaccuracy in reconstructing the target has a direct correlation with the accuracy in the prediction of the expected reward. A lack of target information in the latent representation results in a uniform flat reward prediction.

% When considering the visual target cue 
% The proposed analysis highlights how in the context of visual control, the weight that the pixel information has over the sole state information is predominant. If the target information is provided in a visual manner the system learns to solve the task. When provided as vector information there is a lack of proper encoding of the target information.
 
\textbf{Size matters!} To confirm our hypothesis that the improved predictions are due to the greater significance of the visual targets in the overall loss, we provide additional experiments. In Figure \ref{fig:target_ablation}, we present a study where the Dreamer model is trained on the Reacher environment with varying visual target sizes. We observe that the reduction in pixel information regarding the target adversely affects the target representation within the model, resulting in a deficiency of this information being conveyed to the policy network. The policy struggles to learn to position the entity at the correct location, and we observe that this is correctly reflected in the value function's predictions. This means the policy is aware that is not being able to reach the goal.
With small targets ($<5$ pixels diameters), the representation tends to put more attention on other visually predominant aspects of the environment, struggling to predict the position of the target. In the case of a single pixel target, the amount of target information equals the one of a positional vector and, as expected, the task performance is equally low.

\begin{figure}[!t]
    \centering
    \includegraphics[width=0.9\linewidth]{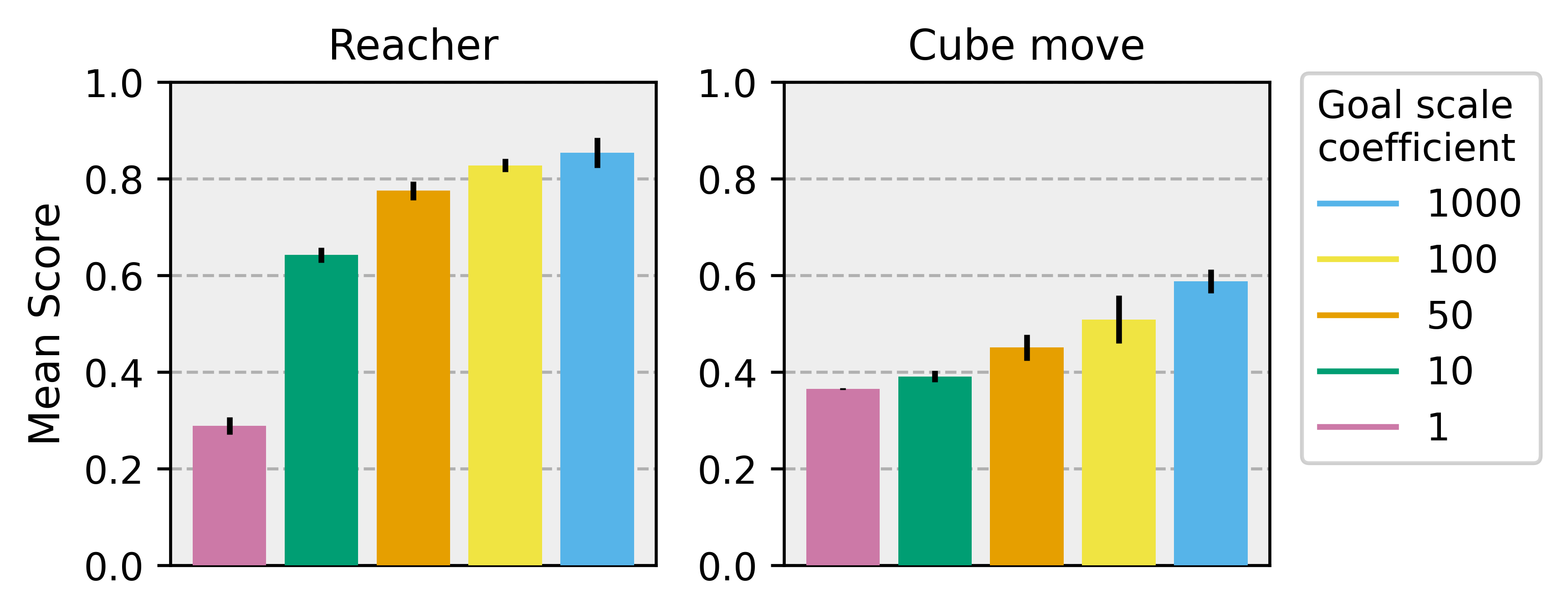}
    \caption{Dreamer trained with goal scaling modulation on the Reacher and Cube move environments.}
    \label{fig:scale_ablation}
\end{figure}

\textbf{Loss rescaling.} To overcome the identified information bottleneck, different strategies can be considered. The simplest one is the re-scaling of the loss components in the decoder to incentivize the model's encoding of the target information. This approach requires finding the optimal scaling factor between the different decoding components, given the complexity of the environment at hand (i.e. 2D or 3D) and the amount of relevant pixels. In Figure \ref{fig:scale_ablation}, we present supporting experiments based on Dreamer, where we vary the importance of the target in the loss of the world model, using different coefficients. We observe that very high coefficients improve the target's reconstruction and thus allow the agent to learn the task. However, the optimal loss coefficient may vary, depending on the complexity of the environment and the presence of information-rich observations. As this naive solution may require extensive hyperparameter tuning for each new scenario, we aim to find more robust strategies for overcoming this issue.

\textbf{Discussion.} Adjusting loss coefficients is a common practice in machine learning.
% Instead, we focus specifically on the  positioning tasks extensively.
A concurrent work \cite{ma2024harmonydreamtaskharmonizationinside} conducted an extensive study between the interplay of the reward and the observation loss in a world model. Our analysis provides an additional insight, as we identify within the observation loss, an unbalance between the different decoded components. In this work, rather than focussing on how to balance the losses, we consider different approaches to alleviate this issue. The central idea is to find alternative ways to provide positional information about the target directly to the reward computation and policy learning modules, rather than relying on the reconstruction of the targets obtained by the model. % and computing reward leveraging, and not rely on the world model to encode it from the inputs to the system. 

\section{Position Conditioned Policy (PCP)}

The first declination of our proposed solutions is the conditioning of the policy directly on the positional coordinates of the desired target. By default, the world model encodes the target's positional information in the latent states, which are then fed to the policy for behavior learning. Instead, as shown in the bottom of Figure \ref{fig:wms}, we propose to concatenate the object positional coordinates $p_g^{obj}$ to the latent states $s_t$ as an input to the policy network $\pi_{PCP}$. 
\newpage
We refer to this strategy as \textit{Position-Conditioned Policy (PCP)}:
$$ \pi_{PCP}(a_t|s_t, p_g^{obj})$$

When employing PCP, the policy has direct access to the target's positional information $p_g^{obj}$. This can also be leveraged for reward computation. Rather than learning a reward head, we can use the world model's decoder to predict the object's position at time t, obtaining $\hat{p}_t^{obj}$. From the experiments in the previous section, we know the agent is generally capable of predicting the object's position. Then, the reward $r_{PCP}$ can be estimated as the distance between the target given to the policy and the reconstructed position of the entity of interest:
\begin{equation}
r_{PCP} = dist(\hat{p}_t^{obj} - p_g^{obj}) 
\end{equation}

This approach requires minimal implementation effort and applies to any world model architecture. We also note that directly conditioning the policy on the goal has been a common practice in model-free RL \cite{Schaul2015UVFA, Andrychowicz2018HER}, which so far has been overlooked for generative world model-based agents.

\section{Latent Conditioned Policy (LCP)}

Conditioning the policy on explicit features has its limitations, particularly when extending features beyond positional ones, or when working with different goal specifications, e.g. visual ones. Therefore, expressing features implicitly could represent a more robust approach. To address this, we propose a latent conditioned method for behavior learning. This approach is analogous to the one adopted in LEXA \cite{mendonca2021discovering} for goal-conditioned behavior learning. However, we tailor our strategy for object manipulation by designing an object-centric approach. We refer to our novel implementation as \textit{Latent-Conditioned Policy (LCP)}.

% world model architecture, based on the FOCUS architecture \cite{ferraro2023focus}. 

\begin{figure}[!t]
    \centering
    \includegraphics[width=\linewidth]{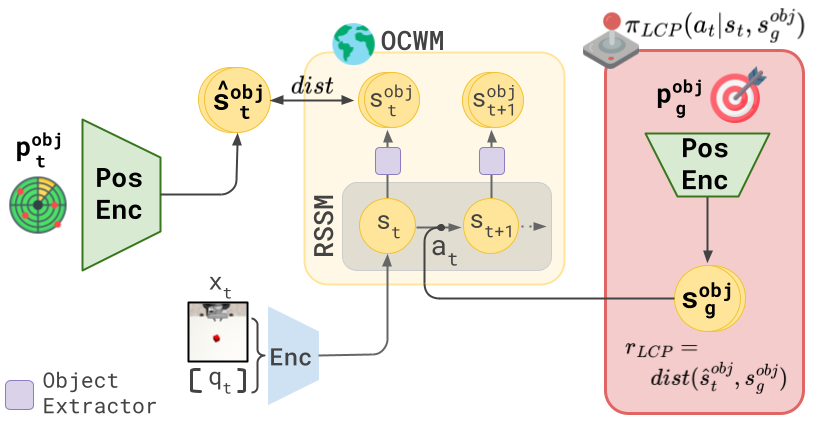}
    \caption{\OurMethod \ leverages an object-centric representation. With the latent position encoder network, the agent learns to predict the latent of each object in the scene given the sole object position. The policy is then conditioned on an object latent target obtained from the target goal observation. Distance functions are expressed as cosine similarities.}
    \label{fig:architecture}
\end{figure}

In LEXA, policy conditioning occurs on the entire (flat) latent state, using either cosine or temporal distance methods. However, in manipulation tasks involving small objects, the cosine approach is inadequate because it prioritizes matching the robot's position over visually smaller aspects of the scene, such as an object's position, rather than on bigger visual components of the scene, e.g. the robot pose. The temporal approach was introduced to mitigate this issue. However, this approach generally requires a larger amount of data to converge, as the training signal is less informative, being based only on the temporal distance from the goal \cite{mendonca2021discovering}.

We argue that object-centric latent representations offer greater flexibility to condition the policy, thanks to the disentangled latent information. With \OurMethod, we can condition the policy solely on the object's latent states, enabling fine-grained target conditioning focused exclusively on the entity of interest.

% We first describe the overall structure and loss of the novel object-centric world model (left in Fig. \ref{fig:architecture}) before delving into more details about the novel policy components of \OurMethod{} (left in Fig. \ref{fig:architecture}).

We now present the novel components with respect to the FOCUS architecture presented earlier. The overall \OurMethod \ method is depicted in Figure \ref{fig:architecture}.
 
% \ref{sec:ocwm}.

\textbf{Latent Positional Encoder}. To obtain object latent features for a given target position, we introduce the Latent Positional Encoder model. This model enables inferring an object's latent state directly from the object's positional information, namely $p(\hat{s}_t^{obj}|p_t^{obj})$. 

During training, the latent positional encoder is trained to minimize the negative cosine distance between the predicted and the reference object latent state: $$\mathcal{L}_{pos} = - \frac{\hat{s}^{obj}_t \cdot s^{obj}_t}{\lVert\hat{s}^{obj}_t\lVert \lVert s^{obj}_t \lVert}$$

Compared to the original loss function of FOCUS (defined in appendix), the world model loss becomes: 
$$\mathcal{L}_{ocwm} = \mathcal{L}_{FOCUS} + \mathcal{L}_{pos}$$

\textbf{Latent-Conditioned Policy Learning}. The introduction of the \textit{latent positional encoder} enables the conditioning over the target object's latent. By encoding a desired target position $p^{obj}_g$, the target object's latent state $s^{obj}_g$ is inferred. The latter serves as the conditioning factor for the policy network $\pi_{LCP}$:
$$ \pi_{LCP}(a_t|s_t, s_g^{obj})$$

To incentivize the policy to move the entity of interest to the target location, we maximize the negative latent distance between $\hat{s}^{obj}_t$ and $s^{obj}_g$. The distance function used is cosine similarity. $r_{LCP}$ becomes then:
\begin{equation}
r_{LCP} = \frac{\hat{s}^{obj}_t \cdot s^{obj}_g}{\lVert\hat{s}^{obj}_t\lVert \lVert s^{obj}_g \lVert}    
\end{equation}
Thus, similarly to PCP, LCP requires no reward head for object positioning tasks.
The latent dynamical consistency of the RSSM allows for the policy network to be trained purely on imagination. 

\textbf{Visual targets.} Additionally with respect to PCP, and similarly to LEXA, LCP enables conditioning the policy on visual targets. In this case, the agent does not use the latent position encoder. Instead, given a visual observation representing the goal target position for the object, the world model can infer the corresponding world model state, using the encoder and the posterior. Then given such a state, the object extractor allows extracting the target latent state $s^{obj}_g$, which is used in the reward computation.

% During the policy network updates targets pose are chosen randomly, according to a uniform distribution, over the entire workspace.   

\begin{figure*}[!tb]
    \centering
    \begin{minipage}{.45\textwidth}
    \centering
    \includegraphics[width=\linewidth]{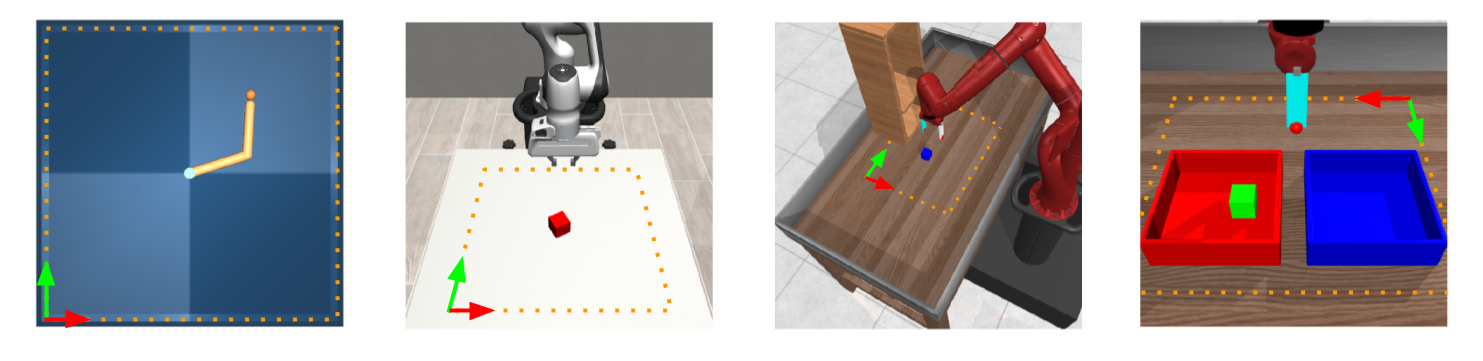}
    % \caption{Manipulation environments: (from left to right) Reacher (DMControl), Cube move (Robosuite), Shelf place (MetaWorld), Pick\&Place (MetaWorld). Workspace for each environment is delimited by the orange dotted line. Reference frames are shown.}
    % \label{fig:envs}
    \end{minipage}%
    \begin{minipage}{0.55\textwidth}
    \centering
    \resizebox{\linewidth}{!}{
    % \begin{tabular}{l|llllll}
%               % &\multicolumn{2}{l|}{}         & \multicolumn{1}{l|}{Position conditioning}& \multicolumn{3}{c}{Latent conditioning} \\
%               & \makecell{Dreamer}       &  \makecell{FOCUS}       & \makecell{Dreamer \\ + PCP } &  \makecell{ FOCUS \\ + PCP }   & \makecell{ FOCUS \\ + LCP }          \\ \hline
% Reacher       & 0.27 ± 0.11   & 0.29 ± 0.1   & 0.8 ± 0.08       &      0.87 ± 0.04 & \textbf{0.91 ± 0.02}           \\ \hline
% Cube move     & 0.35 ± 0.05   & 0.35 ± 0.08   & 0.54 ± 0.04       & \textbf{0.74 ± 0.04} & 0.61 ± 0.05  \\ \hline
% Shelf place   & 0.4 ± 0.06    & 0.3 ± 0.1    & 0.58 ± 0.08       & 0.59 ± 0.1 & \textbf{0.65 ± 0.08} \\ \hline
% Pick\&Place   & 0.26 ± 0.13   & 0.22 ± 0.12   & \textbf{0.48 ± 0.15}   & \textbf{0.47 ± 0.17} & \textbf{0.45 ± 0.17} \\ \hlineB{2.7}
% Overall       & 0.32 ± 0.08    & 0.29 ± 0.09   & 0.6 ± 0.09        & \textbf{0.67 ± 0.09}    & \textbf{0.65 ± 0.08}  \\
% \end{tabular}

\begin{tabular}{l|cccccc}
              % &\multicolumn{2}{l|}{}         & \multicolumn{1}{l|}{Position conditioning}& \multicolumn{3}{c}{Latent conditioning} \\
              & \makecell{Dreamer}       &  \makecell{FOCUS}       & \makecell{Dreamer \\ + PCP } &  \makecell{ FOCUS \\ + PCP }   & \makecell{ FOCUS \\ + LCP }          \\ \hline
Reacher       & 0.27 ± 0.11   & 0.29 ± 0.1   & 0.8 ± 0.08       &      0.87 ± 0.04 & \textbf{ 0.91 ± 0.02}           \\ \hline
Cube move     & 0.35 ± 0.05   & 0.35 ± 0.08   & 0.54 ± 0.04       &  \textbf{0.75 ± 0.04} & \textbf{ 0.73 ± 0.05}  \\ \hline
Shelf place   & 0.4 ± 0.06    & 0.3 ± 0.1    & 0.58 ± 0.08       & 0.59 ± 0.1 &  \textbf{0.70 ± 0.08} \\ \hline
Pick\&Place   & 0.26 ± 0.13   & 0.22 ± 0.12   &  \textbf{0.48 ± 0.15}   &  \textbf{0.48 ± 0.17} &  \textbf{0.43 ± 0.18} \\ \hlineB{2.7}
Overall       & 0.32 ± 0.08    & 0.29 ± 0.09   & 0.6 ± 0.09        &   \textbf{0.67 ± 0.09}    &  \textbf{0.7 ± 0.08}  \\
\end{tabular}
    }
    % \captionof{table}{Average score for 100 goal points equally distributed over the workspace. Performance averaged over 3 seeds, ± indicates the standard error.}
    \end{minipage}
    \captionof{table}{Average score for 100 goal points equally distributed over the workspace. For each task, the environments are shown, in order, on the left. We also show the goal points' workspace, delimited by an orange dotted line, and the reference frames indicated with arrows. Performance is averaged over 3 seeds, ± indicates the standard error. }
    \label{tab:results}
\end{figure*}

\section{Baselines and Environments}

% \begin{figure}[b]
%     \centering
%     \includegraphics[width=\linewidth]{figures/envs_rf.png}
%     \caption{Manipulation environments: (from left to right) Reacher (DMControl), Cube move (Robosuite), Shelf place (MetaWorld), Pick\&Place (MetaWorld). Workspace for each environment is delimited by the orange dotted line. Reference frames are shown.}
%     \label{fig:envs}
% \end{figure}

% \begin{table}[!t]
% \centering
% \resizebox{\linewidth}{!}{
% \input{tables/table_2}
% }
% \caption{Average score for 100 goal points equally distributed over the workspace. Performance averaged over 3 seeds, ± indicates the standard error.}
% \label{tab:results}
% \end{table}

For the evaluation of the proposed method we consider several manipulation environments:
\begin{itemize}
    \item \textbf{Reacher} (DMControl): which, as described previously, represents a pose-reaching positioning task. 
    \item \textbf{Cube move} (Robosuite): where considered target locations are on the 2D plane of the table, no height placement is considered. 
    \item \textbf{Shelf place} and \textbf{Pick\&Place} (Metaworld): The robotic manipulator has to place the cube at the given target location. Considered target locations are on the 2D space in front of the robotic arm.
\end{itemize}

In all environments, the reward signal is defined as the distance between the entity of interest (in the Reacher environment, this is the end-effector) and the target location. All considered environments lack any visual target; the target is provided as an input vector containing spatial coordinates.

We benchmark our methods against various baselines:
\begin{itemize}
\item \textbf{Dreamer}: based on a PyTorch DreamerV2 implementation, but integrated with input vector symlog transformation and KL balancing of the latent dynamic representation, from the DreamerV3 paper.
\item \textbf{FOCUS}: An object-centric world model implementation based on DreamerV2, also integrated with input vector symlog transformation and KL balancing of the latent dynamic representation.
\item \textbf{LEXA}: Based on DreamerV2, this is a latent goal-conditioned method. The conditioning is based on the full latent target. Both proposed distance methods (cosine and temporal) are considered. We adopted our own PyTorch implementation for LEXA.
\end{itemize}

All methods are trained following an offline RL training scheme. The offline datasets contain 1M steps in the environment, which are collected using the object-centric exploration strategy proposed in \cite{ferraro2023focus}. 
% For each agent, the dataset is loaded in a replay buffer and the correspective world models are trained.  
Each model is trained on the dataset for 250K steps, with both the world model and policy network updated at each training step.

\section{Results}

% We present some ablation studies conducted with the intention of highlighting the reasons why Dreamer under perform in this types of tasks. When considering the Reacher environment from the DM Control, it can be noticed how the presence of the visual target is key in the completion of the task. The sole positional information is not enough. Starting from this observation, we design an ablation study to better understand the influence of the visual target in the resolution of this tasks. We run an online setting experiment, in which we variate the radius of the target, without any alteration in the reward function. Results are presented in Fig. \ref{fig:target_ablation}. Decreasing the ratio of target information (e.i. target pixels) have a direct impact on the performance of the agent, caused by the diminishing target-relevant information reaching the actor-critic network. 

We now present the evaluation of the trained models. The main metric considered is the score function presented in Equation \ref{eq:dist}. 
With the experiments, we want to verify whether the introduced techniques, PCP and LCP, overcome the issues presented with object positioning tasks for generating world model-based agents.

First, we verify the performance achieved by the methods when adopting spatial coordinates for the goal definition. We benchmark Dreamer with PCP and FOCUS both with PCP and LCP, and we compare them with the standard Dreamer and FOCUS agents.

Then, we study the FOCUS + LCP method more in detail, by analyzing its multimodal goal specification capabilities. We show a set of experiments where the goal is provided in a visual fashion, using observations with the object positioned at the target location. Here, we compare FOCUS + LCP against the two implementations of LEXA, cosine and temporal.

\begin{figure}[!b]
    \centering
    \includegraphics[width=\linewidth]{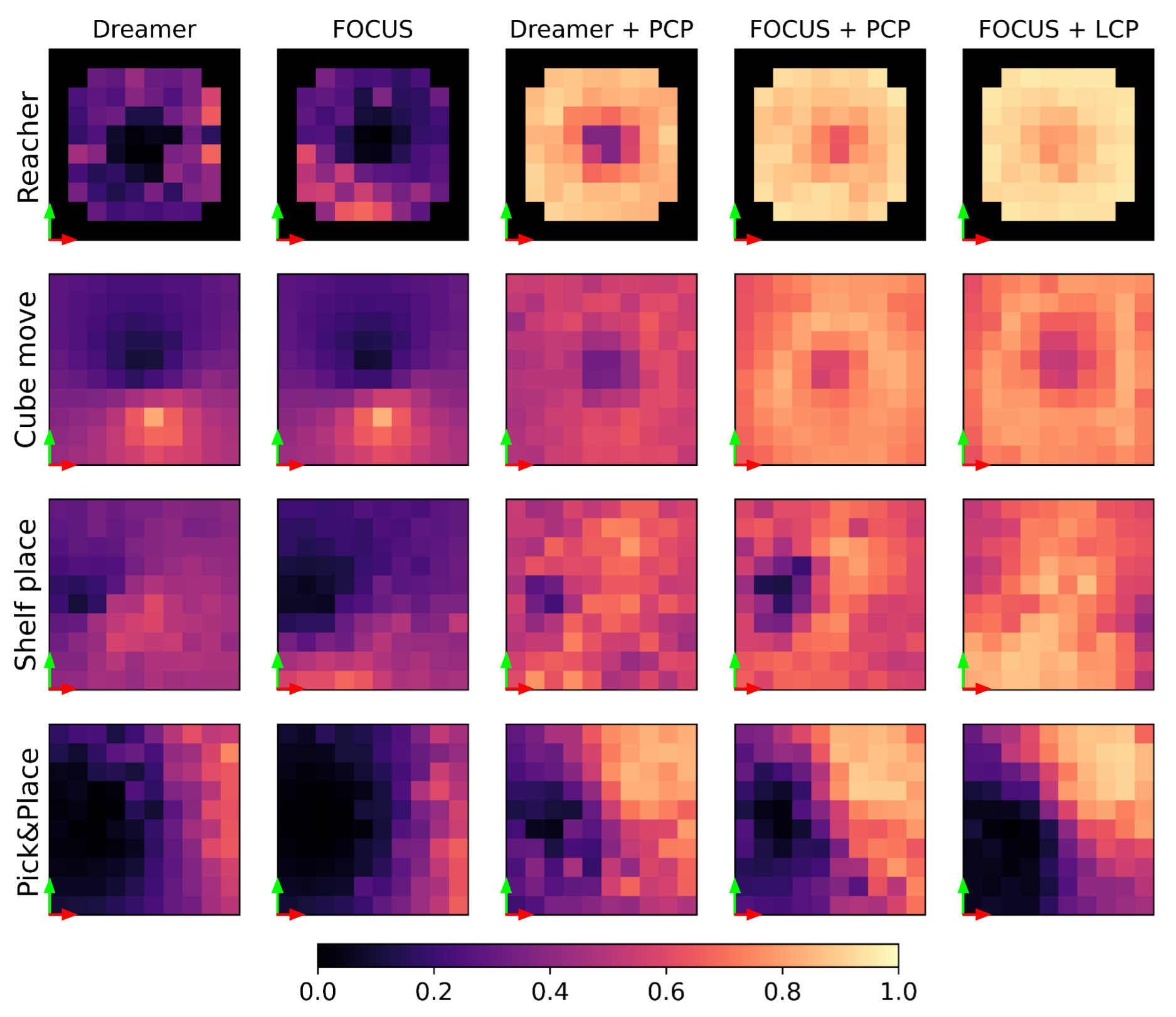}
    \caption{Heatmaps of the mean achieved score for uniformly spread targets in the workspace. References frames refers to the one presented in the figures of Table \ref{tab:results}. The score notation is expressed as the notation presented in Eq. \ref{eq:score_fn}. Results are averaged over 3 seeds.}
    \label{fig:heatmaps}
\end{figure}

\subsection{Spatial-coordinates goal specification}

By providing the different agents with goals uniformly distributed in the workspace we extract the overall performance of each method. 
% For all environments, the evaluation workspace is the same as presented in Figure \ref{fig:envs}. 
Results are presented in Table \ref{tab:results}. 

Overall, the FOCUS agent equipped with PCP or LCP gives the best performance, followed by Dreamer + PCP. In the "Shelf place" environment, the latent representation of \OurMethod \ represents best. Given that the camera is further away from the scene, we believe the agent is better able to deal with the inaccuracies that come from the inaccurate position readings (bigger segmentation mask $\rightarrow$ better granularity in position). 

We also observe that the FOCUS + PCP position is superior to Dreamer + PCP. This is a similar observation as in our analysis of the limitations, where FOCUS tended to perform better than Dreamer. The more accurate predictions of the entity position, in object-centric world models, might be the source of the improved performance.

Finally, as expected, the performance of Dreamer and FOCUS is insufficient in all tasks, and much lower than their PCP and LCP counterparts.
% FOCUS + \OurMethod \ have the best performances in the reaching task for Reacher. 

To highlight the performance distribution over the different goals in the environment, in Fig. \ref{fig:heatmaps} we present heatmaps with the score function for each target location in the workspace. Results are presented for all the different tasks. As expected, both Dreamer and FOCUS have poor performances, resulting in only a few positions being reached with a high score. All the proposed methods have a similar distribution, reaching goals spread all over the environment. % In the Pick\&Place environment, the goals on the left side seem harder to achieve, as the methods achieve lower scores, but this is probably due to the training data distribution.

\begin{figure}[!t]
    \centering
    \includegraphics[width=\linewidth]{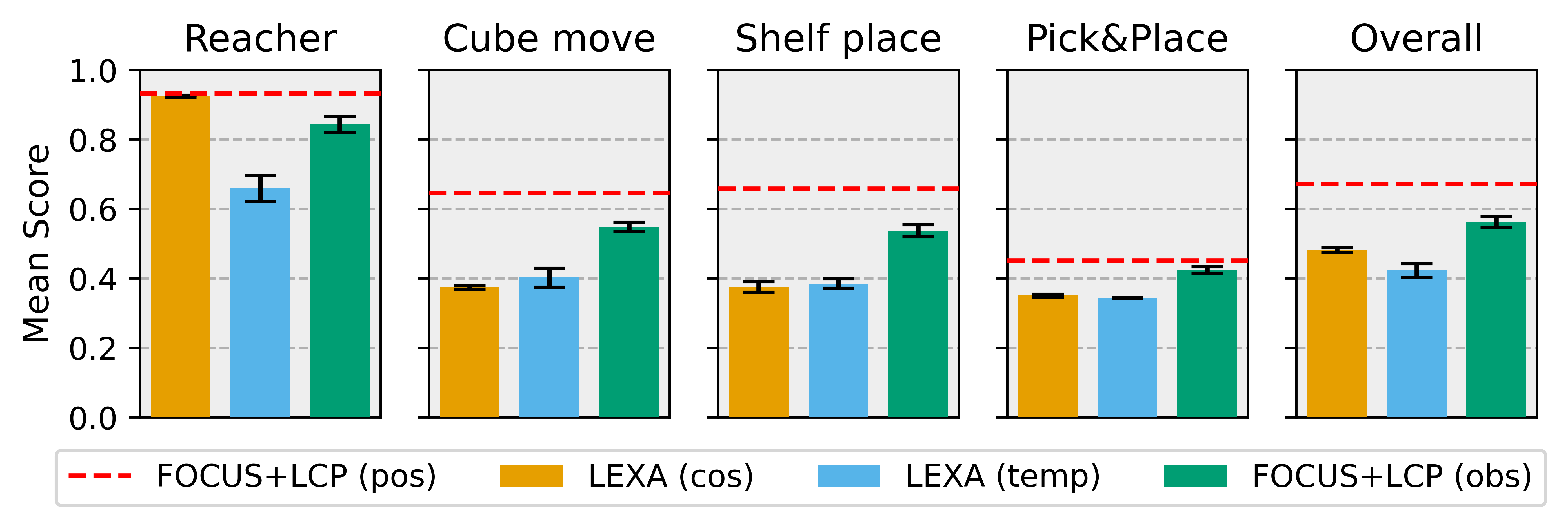}
    \caption{\textbf{Visual goals}. The mean score was achieved over 10 episodes with goal observations for latent conditioning. LEXA is tested both for the cosine and temporal variation. The performance of our method with spatial-coordinate goals (pos) is shown as a reference. The score is expressed according to Equation \ref{eq:score_fn}.}
    \label{fig:multimodal}
\end{figure}

\subsection{Visual goal specification}

An interesting emergence property of FOCUS + \OurMethod \ is the possibility to define goals via different modalities. 
The policy $\pi_{LCP}$ can be conditioned on the goal object latent $\hat{s}_g^{obj}$ coming from the encoding of the visual goal $x_g$.   
% We can now provide a latent goal $s_g^{obj}$ encoding a goal observation $x_g$. The goal observation is encoded and the relevant object latent goal $s_g^{obj}$ is extracted.  
LEXA \cite{mendonca2021discovering} also adopts visual goals for conditioning the action policy, but allows no fine-grained specification of the goal. Thus, to match the goal the agent should accomplish both the agent's pose and the objects' positions from the visual goal. This makes the task harder to accomplish than for our method, where the agent can simply focus on the objects' positions. LEXA temporal is designed to overcome this issue, but it generally requires more data to converge \cite{mendonca2021discovering}.

We compare our method with visual goal conditioning against LEXA cosine and temporal. The goal locations are provided to the simulator which renders the corresponding goal observations by "teleporting" the object to the correct location. The agent is then asked to matched the visual goal, after resetting the environment. Results are shown in Fig. \ref{fig:multimodal}, where the positional conditioning results are shown for reference.    

 As stated before, LEXA matches the flat latent vector to the goal one. This proves helpful in the Reacher environment, where the only part that moves is the agent, and thus LEXA cosine achieves the best performance.
 LEXA cosine fails in the other tasks, given the presence of multiple entities in the observations and visual goals, i.e. the robotic arm and the object. where the model focuses on matching the visually predominant features i.e. the robotic arm. 
 
 LEXA temporal is found to generally underperform. We hypothesize this is due to the need for a larger training dataset, as the training signal, the temporal distance, is less informative than the latent distance, and thus requires more samples from the environment. 

FOCUS+LCP performs better than both LEXA with cosine and temporal distance in all environments but the Reacher. When compared to the performance of FOCUS+LCP with spatial-coordinates goal specification, there is a decrease of only $\sim$10\% in performance. This shows that the approach is promising as a multimodal goal specification method.

\section{Conclusion}

We analyzed the challenges in solving visual robotic positional tasks using generative world model-based agents. We found these systems suffer from information bottleneck issues when considering positional information for task resolution. Under-represented information is challenging for the model to encode, especially in the case of not stationary values such as a variable target. 

The approaches we presented overcome this issue by providing the policy network with more direct access to the target information.
Two declinations are proposed. The first, the Positional Conditioning Policy (PCP), allows direct conditioning on the target spatial coordinates. We showed PCP improves performance for any class of world models, including Dreamer-like "flat" world models and FOCUS-like object-centric world models. The second declination, Latent Conditioning Policy (LCP), is an object-centric approach that we implement on top of FOCUS. This allows the conditioning of the policy on object-centric latent targets.

As future work, it would be interesting to analyze the application of such approaches to non-positional features, such as a shape (e.g. object deformation) or object configuration (e.g. open/close faucet, light on/off). Moreover, it would be interesting to scale the application of LCP to other modalities. For instance, one could consider multimodal inputs including both visual and tactile observations.

% \section{Related Work}

% \textbf{World models.} 
% Numerous methods have been developed to learn visual representations for model-based approaches through image reconstruction \cite{watter2015embed, Seo2022MaskWM,  hafner2019planet, Hafner2020DreamerV1}. Model-based approaches have been proven efficient in solving video games \cite{} and visual robot control tasks \cite{}. 

% A major contribution in the field comes from the work of Hafner et al. \cite{hafner2019planet, Hafner2020DreamerV1, Hafner2021DreamerV2, Hafner2023DreamerV3}, who achieved state-of-the-art performances in different robotics scenarios, thanks to the robust latent dynamic modelling.

\section{Acknowledgments}
This research received funding from the Flemish Government (AI Research Program). Pietro Mazzaglia is funded by a Ph.D. grant of the Flanders Research Foundation (FWO).

\bibliography{references}

\newpage

\appendix

\section{Supplementary Material}

\section{Normalized score}

Scaling performance using expert performance is a common evaluation strategy in RL \cite{cobbe2020procgen, fan2023atarireview}. In our problem, we define the reward as the negative distance:
\begin{equation}
r_t = -r(p^{obj}_t) = -\lVert{p^{obj}_t - p^{obj}_g}\lVert_2.
\label{eq:dist}
\end{equation}

For a given goal $p^{obj}_g$, $r_t \in ]-\inf, 0]$. In order to compare different tasks, where distances may have different magnitudes, we divide the rewards $r_t$ by the typical reward range. This is given by $r_{max} - r_{min}$, where $r_{min} = r(p^{obj}_0)$, with $p_0$ being the initial position of the object (this is normally around the origin, and $r_{max} = r(p^{obj}_g) = 0$.

Thus, we obtain:
\begin{align}
s_t &= r_t / (r_{max} - r_{min}) \\
& = r(p^{obj}_t) / (0 - r(p^{obj}_0)) = \\
& = -\lVert{p^{obj}_t - p^{obj}_g}\lVert_2 / (0 + \lVert{0 - p^{obj}_g}\lVert_2) \\
& = -\lVert{p^{obj}_t - p^{obj}_g}\lVert_2 / \lVert{p^{obj}_g}\lVert_2
\end{align}

Finally, we apply the $\exp$ operator, to make values positive and bring them in the $[0,1]$ range, where 1 is the expert score:
\begin{equation}
\textrm{\texttt{normalized score}} = \exp{ \bigg( - \frac{\lVert{p^{obj}_t - p^{obj}_g}\lVert_2}{\lVert{p^{obj}_g}\lVert_2 }\bigg) }
\end{equation}

\section{FOCUS objective}

Training of the FOCUS architecture is guided by the following loss function:
\begin{equation}
\mathcal{L_\textrm{FOCUS}} = \mathcal{L}_\textrm{dyn} + \mathcal{L}_\textrm{state} + \mathcal{L}_\textrm{obj}.
\end{equation}
$\mathcal{L}_\textrm{dyn}$ refers to the dynamic component of the RSSM, and equals too:
\begin{equation}
    \mathcal{L_\textrm{dyn}} = D_\textrm{KL}[p_{\phi}(s_{t+1}|s_t, a_t, e_{t+1}) || p_{\phi}(s_{t+1}|s_t, a_t)].
\end{equation}
the backpropagation is balanced and clipped below 1 nat as in DreamerV3 \cite{Hafner2023DreamerV3}.

The object loss component is instantiated as the composition of NLL over the mask and RGB mask reconstructions:
\begin{equation}
\begin{split}
    \mathcal{L_\textrm{obj}} = -\log \underbrace{p(\hat{m_t})}_\textrm{mask}  -\log \sum^{N}_{\textrm{obj}=0} \underbrace{m_t^\textrm{obj} p_{\theta}(\hat{x_t}^\textrm{obj}|s_t^\textrm{obj})}_\textrm{masked reconstruction}
\end{split}
\end{equation}

Finally, the decoder learns to reconstruct the rest of vector state information $v_t$ by minimization of the negative log-likelihood (NLL) loss:
\begin{equation}
\begin{split}
    \mathcal{L}_\textrm{state} = -\log p_\theta(\hat{q}_t, p_g^{obj}|s_t)
\end{split}
\end{equation}

\section{Training details and Hyperparameters}
All models in the presented work have been trained offline. Datasets have been collected beforehand, guided by the exploration agent of choice (we tested both the Object-centric entropy maximization proposed in FOCUS \cite{ferraro2023focus} and Plan2Explore \cite{sekar2020planning}). The datasets are loaded in the replay buffer of the offline agents, and the training is conducted for 250K steps. Both world model and agent are updated at every training step. V100-16GB GPUs have been used for all experiments. Our proposed methods (i.e. Dreamer/FOCUS + PCP, FOCUS + LCP) took roughly 18 hours to complete each training run.

The hyperparameters used for the main implementation of the world models and agent are the same used in DreamerV2 \cite{Hafner2021DreamerV2} official implementation. Symlog function is applied at every input. KL balancing as in DreamerV3 \cite{Hafner2023DreamerV3} is implemented.

With reference to FOCUS model, we have the following additional parameters:
\begin{itemize}
\item Object-extractor: MLP composed of 2 layers, 512 units, ReLU activation;
\end{itemize}

With reference to FOCUS + LCP model, we have the following additional parameters:
\begin{itemize}
\item Object-encoder: MLP composed of 4 layers, 400 units, ReLU activation;
\item Distance method object-encoder objective: Cosine similarity (also tested MSE)
\item Distance method actor policy objective: Cosine similarity (also tested MSE)
\end{itemize}

\section{Offline Training Curves}

\begin{figure}[!hb]
    \centering
    \includegraphics[width=\linewidth]{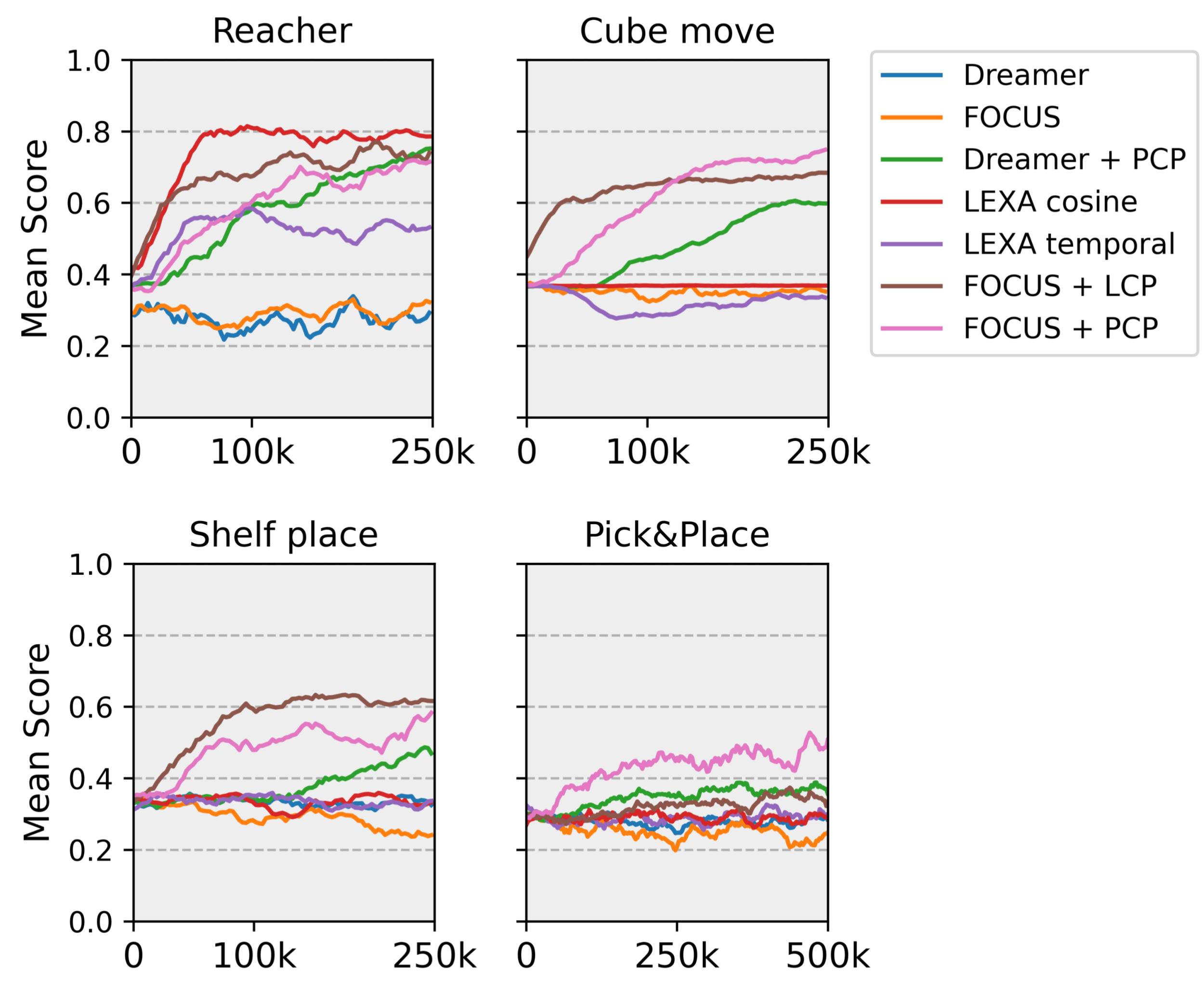}
    \caption{Offline training curves. Standard deviation is omitted for graphical reasons. Mean score refers to eq. \ref{eq:score_fn}} and is computed over 5 evaluation episodes, performed during the offline training. For each episode, a random goal is selected out of a pool of 10 manually engineered ones.   
    \label{fig:training_curves}
\end{figure}

Offline training curves are presented in Figure \ref{fig:training_curves}. In general FOCUS + PCP/LCP have faster convergence when compared to all other methods. Only for the Reacher environment, LEXA cosine converge faster. 

\section{Explorations strategies}

\begin{figure}[h]
    \centering
    \includegraphics[width=\linewidth]{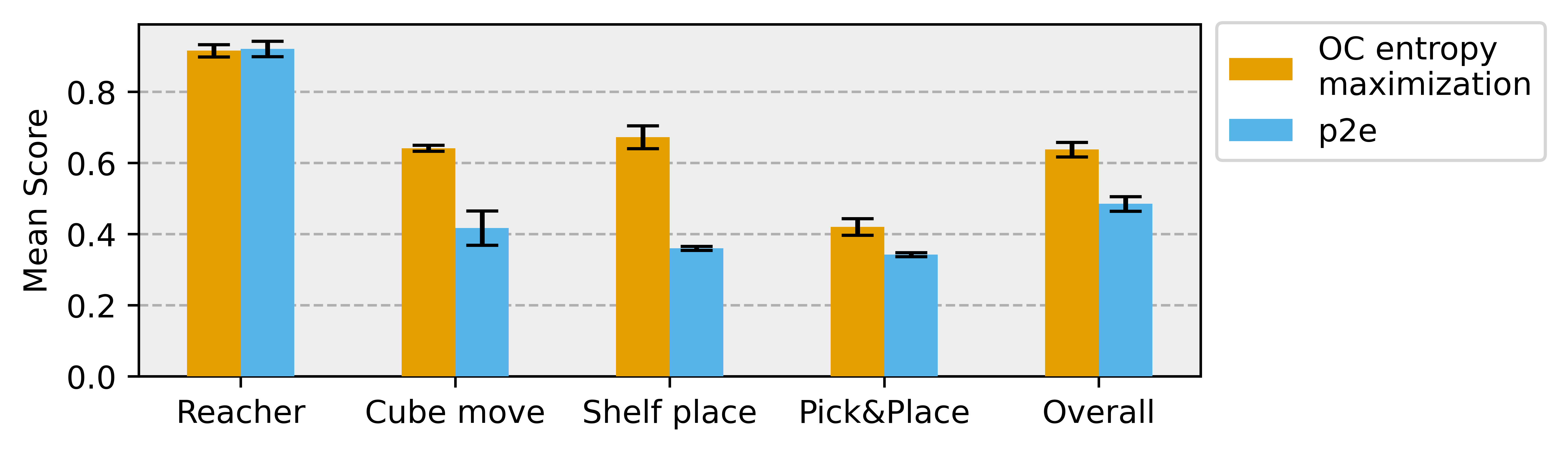}
    \caption{Mean score achieved over 10 episodes for models trained with both datasets obtained from FOCUS exploration method (Object-Centric entropy maximization) and Plan2Explore. The score is expressed according to equation \ref{eq:score_fn}.}
    \label{fig:exploration}
\end{figure}

In the presented work each model is trained offline from a pre-recorded dataset. The dataset of choice is obtained from pure exploration behavior. In Fig. \ref{fig:exploration} we compare the general performance of \OurMethod \ when trained on datasets acquired using different exploration strategies. We consider the object-centric entropy maximization method proposed by Ferraro et al. \cite{ferraro2023focus} and Plan2Explore \cite{sekar2020planning}.  

Overall exploring by maximizing the entropy over the object's latent, gives better performance in the downstream task. We hypothesize this is related to the focus the exploration strategy puts on the object of interest while disregarding background aspects in the scene.  

\section{Reproducibility Checklist}

This paper:
\begin{itemize}
    \item Includes a conceptual outline and/or pseudocode description of AI methods introduced (yes)
    \item Clearly delineates statements that are opinions, hypothesis, and speculation from objective facts and results (yes)
    \item Provides well marked pedagogical references for less-familiare readers to gain background necessary to replicate the paper (yes)
\end{itemize}
Does this paper make theoretical contributions? (no) \\
Does this paper rely on one or more datasets? (no) \\
Does this paper include computational experiments? (yes) \\
If yes, please complete the list below.
\begin{itemize}
    \item Any code required for pre-processing data is included in the appendix. (yes).
    \item All source code required for conducting and analyzing the experiments is included in a code appendix. (yes)
    \item All source code required for conducting and analyzing the experiments will be made publicly available upon publication of the paper with a license that allows free usage for research purposes. (yes)
    \item All source code implementing new methods have comments detailing the implementation, with references to the paper where each step comes from (partial)
    \item If an algorithm depends on randomness, then the method used for setting seeds is described in a way sufficient to allow replication of results. (yes)
    \item This paper specifies the computing infrastructure used for running experiments (hardware and software), including GPU/CPU models; amount of memory; operating system; names and versions of relevant software libraries and frameworks. (yes)
    \item This paper formally describes evaluation metrics used and explains the motivation for choosing these metrics. (yes)
    \item This paper states the number of algorithm runs used to compute each reported result. (yes)
    \item Analysis of experiments goes beyond single-dimensional summaries of performance (e.g., average; median) to include measures of variation, confidence, or other distributional information. (yes)
    \item The significance of any improvement or decrease in performance is judged using appropriate statistical tests (e.g., Wilcoxon signed-rank). (partial)
    \item This paper lists all final (hyper-)parameters used for each model/algorithm in the paper’s experiments. (yes)
    \item This paper states the number and range of values tried per (hyper-) parameter during development of the paper, along with the criterion used for selecting the final parameter setting. (NA)
\end{itemize}

\end{document}